\documentclass{article}

\usepackage{microtype}
\usepackage{graphicx}
\usepackage{subfigure}
\usepackage{booktabs} 

\usepackage[hyphens]{url}
\usepackage{hyperref}
\usepackage[hyphenbreaks]{breakurl}

\usepackage[accepted]{icml2024}

\usepackage{amsmath}
\usepackage{amssymb}
\usepackage{mathtools}
\usepackage{amsthm}

\usepackage[capitalize,noabbrev]{cleveref}

\theoremstyle{plain}
\newtheorem{theorem}{Theorem}[section]

\theoremstyle{definition}
\newtheorem{definition}[theorem]{Definition}

\theoremstyle{remark}

\usepackage[textsize=tiny]{todonotes}

\icmltitlerunning{Position: Embracing Negative Results in Machine Learning}

\begin{document}

\twocolumn[
\icmltitle{Position: Embracing Negative Results in Machine Learning}



\icmlsetsymbol{equal}{*}

\begin{icmlauthorlist}
\icmlauthor{Florian Karl}{iis,lmu,mcml}
\icmlauthor{Lukas Malte Kemeter}{iis}
\icmlauthor{Gabriel Dax}{iis}
\icmlauthor{Paulina Sierak}{iis}
\end{icmlauthorlist}

\icmlaffiliation{lmu}{Ludwig-Maximilians-Universität München, Munich, Germany}
\icmlaffiliation{iis}{Fraunhofer Institute for Integrated Circuits IIS, Fraunhofer IIS, Nuremberg, Germany}
\icmlaffiliation{mcml}{Munich Center for Machine Learning, Munich, Germany}

\icmlcorrespondingauthor{Florian Karl}{florian.karl@iis.fraunhofer.de}

\icmlkeywords{Machine Learning, ICML, Negative Results}

\vskip 0.3in
]



\printAffiliationsAndNotice{}  

\begin{abstract}
Publications proposing novel machine learning methods are often primarily rated by exhibited predictive performance on selected problems. In this position paper we argue that predictive performance alone is not a good indicator for the worth of a publication. Using it as such even fosters problems like inefficiencies of the machine learning research community as a whole and setting wrong incentives for researchers. We therefore put out a call for the publication of ``negative'' results, which can help alleviate some of these problems and improve the scientific output of the machine learning research community. To substantiate our position, we present the advantages of publishing negative results and provide concrete measures for the community to move towards a paradigm where their publication is normalized.
\end{abstract}

\textit{Note from the authors:
Have some of our publications been rejected due to lacking competitive results and has this been frustrating at times? Yes. 
However, the following position paper is not a personal vendetta: we truly believe embracing negative results can be an asset for the machine learning research community and want to present an objective deliberation on why.
We hope to convince you, the reader, of the same in the following pages and spark discussion as well as change in our community.}


\section{Introduction}
\label{sec:introduction}

Machine learning has grown into a prominent research field that has demonstrated large impact on a lot of application domains.
The number of machine learning publications has grown exponentially along with the number of active researchers and funding volume~\cite{maslej-stanfordai2023,krenn-nature2023}. There are many machine learning publications that provide value for the research community: works centered around theory and proofs, benchmarks, survey papers and position papers. However, a large number of machine learning publications examine a (often novel) method and then demonstrate its performance on relevant problems; these are the types of publications we focus on in this work. 

Machine learning is largely an empirical science: If something works and demonstrates good performance it is often deemed a good result and worthy of publication.
On the other hand, if a new method or algorithm is not able to beat the state-of-the-art on a typical benchmark dataset, researchers might quickly abandon their work as it is unlikely to be published.
Despite being a somewhat confusing term when it comes to scientific results, such outcomes are often deemed to be \textit{negative results}.
In science, the terms positive and negative refer to the postulated null hypothesis, which is then either rejected (positive result) or the results of experimentation do not allow for a rejection (negative result). 

\begin{definition}
\textit{The \textbf{usual null hypothesis} of empirical machine learning is that a proposed method does not exhibit significantly better predictive performance than existing methods on a relevant subset of problems.}
\end{definition}

In this terminology there is no room for ``good'' or ``bad'' results.
However, machine learning as an empirical science has developed a strong attachment to predictive performance and it often seems that only very specific positive results, those that show that a proposed method beats the state-of-the-art, are considered ``good'' results.
In the context of this paper, when talking about negative results we refer to the following, specific case:

\begin{definition}
\textit{A \textbf{negative result} in empirical machine learning research occurs, when the usual null hypothesis can not be rejected.}
\end{definition}

We want to distinguish between two important subtypes of negative results: Novel method negative results (NMNR) and existing method negative results (EMNR).
NMNR mostly refer to submissions proposing a novel method that does not beat existing state-of-the-art methods with respect to a suitable performance metric on selected test problems.
EMNR occur, when existing methods, that are considered state-of-the-art, are demonstrated to have inferior performance as to what was expected.
This could be a replication study or a publication shining a light on specific failure modes of existing methods.
A famous example is the work by~\citet{bengio-ieetnn1994}, in which the authors lamented the problem of vanishing gradients when training recurrent neural networks and went on to further explore and analyze these negative results.
EMNR-publications arguably have a better standing in the community and are thus published more often as compared to NMNR-publications, but we will still include them where relevant as they also constitute negative results. 

Similarly, when talking about positive results, we refer to the following:

\begin{definition}
\textit{A \textbf{positive result} in empirical machine learning research occurs when the usual null hypothesis is rejected.}
\end{definition}

Neglecting negative results and almost exclusively publishing positive results leads to a number of problems like publication bias~\cite{boulesteix-cancerinformatics2015}, inefficiencies in the community, a disconnect between machine learning research and application as well as setting problematic incentives for researchers.
\textbf{This position paper argues that machine learning research is at a point where it should encourage or even welcome the publication of negative results.} 
To be clear: We are not advocating for a majority of publications to be centered around negative results, but believe the scientific output of our research community can be healthier and ultimately better if we encourage parts of it to do so.
To this end, we examine the problems introduced into the machine learning research community through an overemphasis on predictive performance in Section~\ref{sec:ml_for_accuracy}.
We specifically discuss the shortcomings of overly relying on predictive performance as a proxy indicator for worth of newly proposed methods, inefficiencies in the machine learning research community and unhealthy incentives for machine learning researchers.
In Section~\ref{sec:embracing_negative_results} we highlight the positive effects that could result from normalizing the publication of negative results.
To further stimulate discussion we then propose some counterfactual arguments in Section~\ref{sec:counterfactuals} before giving some concrete recommendations on how to move forward in Section~\ref{sec:how_to_do_it}.
In this Section we also highlight some developments and previously taken measures that already help push the machine learning research community in the direction we propose in this work.

\section{Related Work}
\label{sec:related_work}
The topic of negative results is not unknown to science and society in general. 
Specialized journals in other research fields like the \textit{Journal of Negative Results in Ecology and Evolutionary Biology} or the \textit{Journal of Pharmaceutical Negative Results} demonstrate the topic's importance to a number of other research fields. 
Research within the field of economics even considers the effects of publishing negative results (for science in general) from a game theoretic perspective.
\citet{bobtcheff2021} e.g., model the problem of publishing negative results as a competition between players. 
They identify conditions for which publishing negative results is preferred both by society as well as the players (i.e., competing researchers): This occurs in circumstances that are not winner-takes-all settings (like a race for filing the first patent)---meaning when competition is not too fierce. 

The notion of publication bias is closely related to negative results.
Publication bias is a known quantity in science and has long been studied~\cite{sterling1959publication,boulesteix-cancerinformatics2015,ritchie2021science}; perhaps most famously in medical and clinical science~\cite{easterbrook1991publication}.
Publication bias in science in general is linked to an increase in false positive research findings in published results. 
False positives in this context refer to findings that are mistakenly identified as significant or meaningful when, in reality, they are not. 
\citet{boulesteix-cancerinformatics2015} is the first work to formalize the notion of publication bias for methodological computational research. This bias is reinforced by gatekeeping negative results and setting incentives for researchers to produce positive results. 
While the topic is proposed in the context of cancer informatics, the presented ideas and concepts hold true for machine learning research in general.

We believe that some unique properties of machine learning research make embracing negative results an especially important topic for our research community. We will discuss these in  Section~\ref{sec:ml_for_accuracy}. Some previous works discuss the topic of negative results in machine learning research specifically.
\citet{boulesteix-cancerinformatics2015} provides a pilot study about publication bias and negative results in methodological computational research and includes some discussion on the effects of publication bias on this research area.
The introduction by~\citeauthor{giraud-sigkdd2011} to the \textit{ACM SIGKDD Explorations Special Issue: Unexpected Results} provides deliberation on the importance of negative results, but serves mainly as a prelude for the following articles that make up the special issue.
Additionally, it is quite short and only briefly touches upon some of the benefits that the publication of negative results can foster.
Finally, it has been published fourteen years ago and we believe some of the points raised in the following sections to be especially important now as compared to then, e.g., due to the fast-paced growth of machine learning research over the last decade.

Some selected venues have embraced the paradigm of publishing negative results in the past.
The online journal \textit{Interesting Negative Results in Natural Language Processing and Machine Learning}\footnote{\url{http://jinr.site.uottawa.ca/}} unfortunately did not see a lot of activity in recent years.
The Annual Conference on Neural Information Processing Systems hosts the \textit{I Can't Believe It's Not Better!}\footnote{\url{https://i-cant-believe-its-not-better.github.io/}} workshop, which showcases unexpected negative results in machine learning.
This workshop is still active and saw its most recent edition in 2023 focus on ``Failure Modes in the Age of Foundation Models''.
Finally, the \textit{Workshop on Insights from Negative Results in NLP} attempts to combat a perceived overemphasis on benchmark results in the field of natural language processing (NLP) by inviting researchers to submit ``both practical and theoretical unexpected or negative results that have important implications for future research, highlight methodological issues with existing approaches, and/or point out pervasive misunderstandings or bad practices''\footnote{\url{https://insights-workshop.github.io/}}.
While these are some positive examples, we would like to see the topic of communicating negative results become an established pillar of machine learning research. 

\section{Machine Learning for Accuracy's Sake}
\label{sec:ml_for_accuracy}
Metrics for predictive performance are of central importance to a lot of machine learning publications with accuracy arguably the most famous one among them.
In the following we will formulate three key hypotheses to highlight several issues within the machine learning research community that stem from an overemphasis on predictive performance in publications and reviews alike.

\subsection{Pure Predictive Performance Is a Faulty Metric for Scientific Progress}
\label{ssec:bad_metric}
The machine learning research community has chosen a problematic metric for scientific progress and the worth of publications and a noisy one at that.
To be fair, judging publications (and especially NMNR-publications) and their contribution is not straightforward.
As~\citet{wagstaff-arxiv2012} once put it: ``what is the field’s objective function?''
It is unclear what machine learning should strive towards.
Performance gain? Impact on society?
While many venues give detailed instructions for their reviewers and encourage them to judge aspects such as novelty, significance and relevance~\cite{icml-review-form}, these abstract terms are often hard to gauge compared to an improvement in predictive performance.

Comparing our field of research to e.g., medicine: the evaluation of (novel) procedures and medication is clearer than that of machine learning algorithms.
Although studies may be conducted with too few participants, results may be misinterpreted etc., the metric for success is above reproach. 
If a medication helps more people become healthy more quickly than existing medications, it can be deemed a success and can be believed to have a positive impact on society.
In contrast, if a novel computer vision method can increase classification accuracy on popular benchmark sets like ImageNet or CIFAR, it cannot be said with any certainty that this method will help practitioners solve the problems they face every day or have an impact on society otherwise.
This lack of connection between research and actual applications of machine learning has long been lamented within the community~\cite{roberts-nature2021,wagstaff-arxiv2012,liao-nips2021, varoquaux2022}.
Take the recent example of~\citet{roberts-nature2021}.
They examined machine learning models proposed in publications to detect and/or prognosticate coronavirus disease with regard to potential clinical use.
They identified 2,212 relevant studies and scrutinized the most promising 62 studies after extended quality screening. They have found none of the proposed models to be of any clinical use. Similarly, \citet{varoquaux2022} speak of a mismatch between the evaluation approach for practical machine learning applications and research benchmarks in the medical domain. 
They lament dataset bias\footnote{Test sets for benchmarks are often random subsets of the training domain, while in practice, the training data distribution often differs from the application distribution.}, faulty metrics, and improper evaluation procedures among others.

That is not to say that all empirical machine learning publications have an overemphasis on predictive performance.
Often secondary metrics like efficiency, interpretability or robustness to domain shifts are considered and indeed a large portion of real-world machine learning applications have to contend with multiple objectives and need to carefully consider trade-offs between them~\cite{jin2006multi}.
Additionally, some metrics quantifying predictive performance for certain use cases or even whole subfields of machine learning are better suited for evaluation because they are known to (better) map to a real-world problem. 
While this scenario---independent of the considered metric(s)---will always be prone to a certain ``leaderboardism'', this is arguably less of a problem if metrics are aligned with problems practitioners are trying to solve with machine learning.
Unfortunately, this does not apply for a high number of publications.
Among many other factors the choice of metric along with the rest of the empirical evaluation setup is an important factor to consider when judging relevance and impact of empirical machine learning publications.

Furthermore, a lot of the improvements on standard problems are nowadays only minimal.
Machine learning continues to post state-of-the-art results, but year-over-year improvement on many benchmarks continues to be marginal~\cite{maslej-stanfordai2023}. \citet{varoquaux2022} refer to this as ``diminishing returns''. After assessing \textit{Kaggle} competition results, they found that in multiple cases, the reported performance gains by the winners were smaller than the evaluation noise, meaning no actual improvement was achieved.
Research is reaching performance saturation on several traditional benchmarks, so this is to be expected to some degree~\cite{maslej-stanfordai2023}.
On the other hand, predictive performance is sensitive to e.g., cherry-picking datasets~\cite{balduzzi-nips2018}, tuning hyperparameters~\cite{yang-iclr2020}, tricks in the evaluation protocol~\cite{yang-iclr2020} or even random seeds~\cite{picard-arxiv2021}. 
The community therefore increasingly struggles with evaluation of newly proposed methods and has developed a distrust of these minimal improvements.
Indeed, several publications in the past have pointed out growing issues with reproducibility of published results and have called the current state of machine learning research a \textit{reproducibility crisis}~\cite{kapoor-patterns2023,pineau-nips2021i}.


\subsection{A Hyper-Focus on Predictive Performance Sets Bad Incentives for Researchers}

If submissions that demonstrate improved predictive performance through a novel method are overly rewarded in the review process, this sets certain incentives for researchers.
E.g., machine learning research could benefit immensely from researchers re-implementing existing methods, benchmarking them and publishing their results.
But we see such publications rarely, because there is no incentive to write them: Researchers could just as well propose a novel method and have a much higher chance of publication.
Furthermore, computing resources have become integral for improving the state-of-the-art in several subfields of machine learning, like e.g., generative artificial intelligence. 
Indeed, a majority of significant machine learning models were produced by industry as compared to academia~\cite{maslej-stanfordai2023}.
By overly rewarding performance improvements and beating the state-of-the-art, the community essentially makes resource inequality a gatekeeper to publications and only allows a selected few to shape important parts of the research field.
While not all machine learning papers are empirical (see Section~\ref{sec:introduction}) and there are many ways to get published without an abundance of computing resources, the availability of computing resources is in our opinion an important factor.
Computing resources are for similar reasons often considered a confounding variable in the context of experimental results; we revisit confounding variables in a broader sense in Section~\ref{sec:counterfactuals}.
Researchers are also encouraged to take less risks. 
In a fast-paced research environment, where many researchers have to regularly publish, people tend to pursue projects, that have a low likelihood of producing negative results.
As comparatively little reward is given for unique ideas that do not demonstrate performance gain over other methods, there is less incentive for innovation and speculative ideas.
Of course, innovative ideas are published (ideas can be unique and have a low probability for negative results) but we conjecture that many interesting ideas are never pursued or published, because our current state of research does not set the right incentives in this respect and thus stifles innovation.

\subsection{Machine Learning Research Has Become Increasingly Inefficient}
\label{ssec:efficiency}
The research field of machine learning has grown at a staggering pace over the past couple of decades. 
The number of publications related to artificial intelligence has more than doubled between 2010 and 2021, reaching an amount of almost 500,000 in total by the end of 2021~\cite{maslej-stanfordai2023}.
\citet{krenn-nature2023} have observed exponential growth in the number of papers published each month with a doubling rate of roughly 23 months surpassing the astonishing number of 4,000 papers per month in 2023.
There is a constant influx of new minds, and a large amount of funding is dedicated to machine learning research~\cite{maslej-stanfordai2023}.
This has led to our community being a fast-paced research environment, which is a boon in many ways.
We have witnessed a great amount of innovation in the last few years alone with large language models and generative artificial intelligence leading the charge recently.

However, the sheer number of people working in machine learning research have made the research community inefficient.
Even in specialized sub-fields people are bound to research the same problems, discover new methods and come to similar conclusions.
This is only a small problem in case of success (i.e., publication of a novel method).
Worst case for the research community is two somewhat similar papers that have a slightly different spin on this method being published around the same time.
A good example for this are GoogLeNet~\cite{szegedy-cvpr2015} and VGG networks~\cite{simonyan-iclr2015}, which were developed independently and in parallel.
However, many methods examined by researchers produce negative results, as is the nature of research.
Without publishing some of these negative results, other researchers may attempt to validate similar methods in similar experiments.
The research community is destined to act akin to a reinforcement learning algorithm without negative feedback.
The inefficiencies extend to allocation of funds as well as computational resources.
Many models are expensive to train and a lot of computing resources are wasted trying things that have already been tried.
Other scientific fields have solved this inefficiency problem by e.g., pre-registering studies or experiments before their conduction so as to avoid these inefficiencies---at least for larger projects~\cite{ritchie2021science}.
Despite recent efforts like the \textit{NeurIPS Workshop on Pre-registration in Machine Learning}, pre-registration has not caught on for machine learning research~\cite{hofman-arxiv2023}.
We believe this to be in part due to its fast-paced nature and the flexibility researchers have to exhibit because of it.



\section{Impact of Embracing Negative Results in Machine Learning}
\label{sec:embracing_negative_results}
If a reviewer finds themselves in front of a paper which proposes a new method, how should they decide if the work is worthy of publication?
According to the guidelines of many conferences it should be judged by significance, relevance and novelty alongside other aspects such as overall soundness, quality or presentation~\cite{icml-review-form}.
Ultimately, it should be judged by its potential impact on and advancement of the research field as well as impact on society. Is this something people will benefit from when they read it?
The problem is not that reviewers do not follow these guidelines, but that performance of a proposed method has become an easily measurable stand-in for this more abstract worth of a paper.
As more and more positive results are published, researchers become more inclined to submit only similar works for review.
Following this spiral, we have achieved a state where thousands of papers are published each month introducing new methods that all seemingly surpass the state-of-the-art.
Publishing NMNR-publications---if they are deemed to likely have a positive and sufficient impact---can break this spiral and re-calibrate how we rate newly proposed methods.
Liberating the research field in this respect could lead people to not pursue the work that will get them published, because it has some small performance gain, but to increasingly pursue the research they deem important for the community (and still get published).

Having interesting and novel ideas published, even if they do not result in a performance improvement, introduces those ideas to the many bright minds that work in machine learning research.
They might themselves have ideas to expand on it or tweak the original proposed method to maybe find success after all.
Understanding why a particular approach did not yield the expected outcome can lead to new insights and improvements in methodologies or theory.

Furthermore, if some interesting ideas with ultimately negative results are published and become a part of the scientific bodies, others will not succumb to the allure of this interesting idea, because they know it will not work. 
And if they still do suspect potential in this idea and want to further pursue work in this direction, they have a much better starting point.
There are plenty of negative results that can help the machine learning research community advance.
Especially EMNR-publications have demonstrated this in the past. Vanishing gradients~\cite{bengio-ieetnn1994} ultimately led to the to the introduction of long short-term memory architectures by~\citet{hochreiter-neuralc1997}, a method that specifically targets and alleviates this weakness in recurrent neural networks.
Another prominent example is adversarial examples, which were first observed and named in~\citet{szegedy-iclr2014}.
Through small perturbations a network can be prompted to misclassify an image, which it would otherwise classify correctly.
Adversarial examples were further considered in~\citet{goodfellow-iclr2015} and have since spawned an active research community around them, which has helped make neural networks more robust and reliable. 

Encouraging the publication of negative results could also lead to an increase in EMNR-publications that meticulously test or reproduce results from previous work. 
This could be a great step towards alleviating the reproducibility crisis mentioned by ~\citet{pineau-nips2021i}.
Another aspect related to reproducibility: Meticulous science is equivalent to documenting all significant results. Registering what does not work is no less important than registering what works. 
Every researcher does this for themselves, subconsciously or purposefully, so it stands to reason that the machine learning community as a whole can profit from this if adopted in a reasonable manner.

Finally, the publication of negative results will foster a more comprehensive and nuanced understanding of our own research field. 
Taking some of the focus away from performance gains may open up room for theory to catch up with empirical results. 
Focusing on a broader notion of impact and not strictly on predictive performance encourages more diverse research as interesting approaches may be rewarded regardless of their performance, which could ultimately lead to a better theoretical foundation.


\section{Counterfactuals}
\label{sec:counterfactuals}
To further stimulate discussion in the community, we want to present several counterfactuals to our central hypothesis.
While we want to highlight the most common counterfactuals, that we discovered during research and in discussion with our colleagues, our answers to them are quite similar.
Almost all the following arguments (counterfactuals 2--4) against normalizing the publishing of negative results can also be made in the context of publications with positive results.
Furthermore, many of the risks outlined in the following counterfactuals can be alleviated by a healthy review process.
A healthy review process is not something that should have to be introduced anew; such a review process is imperative to the way machine learning research functions today: with or without negative results being published.

\textbf{1) Publication of negative results lowers the overall quality of research in the field.}
Without positive, significant findings, papers might lack the rigor or innovation typically expected in published research.
We agree that the average publication with a performance improvement over the state-of-the-art is likely to be more impactful than the average publication without one.
However, relying too heavily on predictive performance in the judgment of newly proposed methods, is akin to a machine learning model making decisions on one noisy feature that has some correlation with the target instead of using all available features to achieve a higher performance\footnote{Yes, there is some irony to this argument, but the machine learning model in this metaphor is used in an application, not proposed as a novel method in a publication.}.
Reviewers should judge submissions based on all ``features'' available to them; that way only high-quality works will get published.
After all, if experimental design was sound, analysis well done and capable of sufficient discrimination to produce confident results, there can also be value in negative results.
Finally, as outlined in Section~\ref{sec:introduction} we do not advocate for a majority of publications to be about negative results; the better part of published works should rightfully not be centered around negative results.

\textbf{2) Knowing a method does not work in a specific setting has limited value. Knowing it does work in a specific setting is inherently of higher value.}
Who is to say negative results only occur, because hyperparameters haven’t been tuned properly, the proposed method is validated on the wrong type of problem or even due to faulty implementation?
However, a lot of these arguments hold true the other way around. Maybe $\epsilon$-improvements are only achieved because of a specific hyperparameter setting or cherry-picked datasets.
We actually believe this is not an issue of negative vs. positive results, but rather one of proper evaluation of methods and meticulous experimental protocol.
These issues can arise in papers presenting negative and positive results alike.
This extends to the more abstract concepts of confounding variables when it comes to empirical results.
Some, like finetuning of hyperparameters or computing resources are more commonly associated with positive results and others, like implementation errors, are often associated with negative results.
If authors do not indicate or demonstrate clearly what exactly contributed to their results through e.g., clarifying comments, understandable and meticulous experiments or ablation studies, reviewers are called upon to address such shortcomings, ask the necessary questions and if required reject such submissions---for both positive and negative results.

\textbf{3) New proxies for scientific worth of publications will emerge and a new bias is introduced into what is published.}
Researchers are bound to optimize their submissions, which may give rise to new proxies in place of performance that are then overvalued. 
One such example would be chasing of trends to pander to current topics.
We argue that these proxies already exist today; overpublishing certain topics, because they are ``trendy’’ is not a new thing.
In our opinion, performance is merely the most prominent one.
Again, we put faith in the review process, which already has to contend with these challenges to also solve these problems in the future.

\textbf{4) Certain types of negative results are more likely to be published than others.}
There's a risk of creating a bias towards publishing only certain types of negative results, potentially those that align with popular narratives or current trends, rather than a truly representative sample of all negative outcomes.
This is also true for positive results and a current problem of machine learning research.
We believe the publication of negative results will not change or intensify this problem.

\section{How to Embrace Negative Results in Machine Learning}
\label{sec:how_to_do_it}
To conclude this position paper, we want to propose some measures that could help pave the way towards a new paradigm in machine learning research where the publication of negative results has been normalized and showcase some efforts that have already been made in this direction.

\subsection{Create special issues, workshops or conference tracks that especially encourage negative results.}
If these results are specifically encouraged at top conferences and journals, people will attempt to submit their negative results.
Such special issues or tracks could further act as a catalyst towards this new mentality of publishing negative results.
Workshops are especially suited for this, as they are intended to stimulate discussion in the machine learning research community.
One of the benefits outlined in Section~\ref{sec:embracing_negative_results} was innovation and exciting research building on top of negative results. 
Workshops, which are dedicated to exchange, are a great platform to fulfill this promise.
The venues and journals mentioned in Section~\ref{sec:related_work} (\textit{I Can't Believe It's Not Better!} workshop, \textit{Interesting Negative Results in Natural Language Processing and Machine Learning} online journal, \textit{Workshop on Insights from Negative Results in NLP}) are a great first step in this direction.
The \textit{ACM SIGKDD Explorations Special Issue: Unexpected Results} that showcased some negative results~\cite{giraud-sigkdd2011} was unfortunately not followed up on.

\subsection{Encourage researchers to discuss negative results in the context of their research, even if this is not the main focus of their publication.} Machine learning venues should encourage submitting researchers to discuss failures and key learnings from their research project even if their method now beats the state-of-the-art.
Those researchers likely learned a lot throughout the project and it can be valuable to share these insights.
We believe this is not something venues should actively incentivize by e.g., rating submissions that include a section like this more highly.
This would unfairly punish people who do not have any interesting learnings to share (yet might still have a great publication) and may even lead to researchers including some ``fake failures''.
We instead suggest venues actively encourage researchers to include such findings and deliberation in their submissions (even if it is very short or in an appendix so as to not take up too much space) and over time evaluate if researchers respond positively.
While not a peer-reviewed publication, \citet{redmon2018yolov3} is an example of an influential publication which includes such a section.
While they do not delve deeply into the analysis of things that did not work and, in the scope of the proposed model, the discussed modification that resulted in surprising negative results are fairly small, this content may nonetheless be interesting to readers and especially researchers from the same subfield.

\subsection{Challenge papers should include failed attempts.}
Some venues propose a challenge before e.g., a conference.
Researchers can submit their solutions as well as results and oftentimes, the winners are asked to prepare a submission detailing their solution.
As the problem and evaluation scheme are set beforehand, a strong emphasis on predictive performance is sensible in this case: The whole point of a challenge is to compete with respect to a set metric on the given problem.
However, we would like to see venues encourage those submissions to also contain a section on what researchers tried before their winning solution and what did not work as well.
There is no pressure on the authors as they are guaranteed to be published through having won the challenge beforehand, which provides a setting in which they can discuss their negative results freely.
One example where this was realized is the \textit{iWildCam challenge 2022}\footnote{\url{https://www.kaggle.com/competitions/iwildcam2022-fgvc9/}} as part of \textit{The Ninth Workshop on Fine-Grained Visual Categorization (FGVC9)} at the \textit{IEEE / CVF Computer Vision and Pattern Recognition Conference (CVPR) 2022}, where participants were explicitly encouraged to share any surprising negative results.

\subsection{Include important negative results in teaching.}
Teaching why things do not work can be as beneficial as teaching why they do.
This reinforces important principles about scientific and critical thinking.
If publishing of negative results should become normalized, education of new researchers is an important place to start teaching the benefits of this.
We believe this to already be implemented by some lecturers in a variety of contexts.
To address two examples we have showcased as impactful EMNR-publications in Section~\ref{sec:embracing_negative_results}:
In several courses on sequential learning and time series forecasting long short-term memory architectures are motivated through negative results obtained from recurrent neural networks in certain applications\footnote{See e.g., Andrew Ng's \textit{Coursera} lecture on sequence models at \url{https://www.coursera.org/learn/nlp-sequence-models}.}.
Similarly, robustness of neural network architectures is often motivated in lectures through adversarial examples.

\subsection{Specifically incentivize replication studies with publications and funding.}
We believe replication studies and validation studies of previously proposed methods to be an especially important subtype of (possibly) negative results that can have great impact on the research community.
While this is probably not feasible for all venues, we hope that some will start accepting these types of publications and maybe actively encourage their submission through special tracks or special issues.
Funding also needs to be dedicated for these types of projects, so people see the worth in pursuing such work.

\subsection{Open subfields of machine learning to embrace negative results.}
We have addressed the rapid growth and overall size of machine learning research in Section~\ref{ssec:efficiency}.
While we would welcome special issues and conference tracks on negative results in machine learning in general, for this to really permeate the community, we believe this has to extend to subfields like time series forecasting, automated machine learning, object detection, etc.
Whether those are application domain-specific or tailored to specific methodology is of lesser importance.
For a paradigm shift to happen, specific measures need to be implemented in ``smaller'' research communities within machine learning, so researchers have a realistic chance of publishing their important negative results.
If there is only one conference workshop a year for all of machine learning, this is simply not feasible.
The \textit{I Can't Believe It's Not Better!} workshop (their 2023 edition e.g., focused on foundation models) and the \textit{Workshop on Insights from Negative Results in NLP}, which targets only NLP research, are good examples and promising first steps.

\subsection{Make a conscious effort to adapt the review process to better accommodate negative results.}
A comprehensive re-design of the review process is unfortunately out of scope for this work. Nevertheless, we did not want to leave the topic untouched. The review process as implemented in most machine learning venues (small variations aside) is certainly not perfect, but has many positive traits and is very important for a healthy scientific community. In the following, we want to outline three ideas of how the review process could be adapted to better accommodate negative results

A very simple measure could be reviewers' guidelines including a small informative section to raise awareness about negative results. This could touch on why a pure focus on predictive performance has downsides as well as explain the different types of negative results, what value negative results can have and what to look for as a reviewer (see Section~\ref{ssec:review-criteria-negative-results}).

A second idea we consider worth discussing is related to the concept of pre-registration, but adapted for the fast-paced environment of our field. 
The review process could be split into two phases. 
Initially, authors are asked to submit a shorter, redacted version of their work, which details the idea, experimental setup etc., but does not mention experimental results. 
Results should then be included in the final version, which is then re-examined for soundness, quality and other applicable criteria by reviewers.
This deliberately eliminates bias from experimental results in the first decision round and re-emphasizes other virtues while still allowing for inclusion of results in the final decision regarding acceptance.
This idea specifically targets NMNR-publications as the initial decision is made independent of the achieved predictive performance, but is also applicable in the context of some EMNR-publications like replication studies.

A last suggestion targeted at NMNR-publications, that to the best of our knowledge has not been discussed, is to adopt a strategy inspired by policy making and regulation. When regulating markets it is sometimes good practice to change which party is responsible for specific actions or for providing relevant information. 
Researchers could be encouraged to submit negative results but be asked to provide an additional small deliberation already on submission (ex ante) explaining why the paper provides value for the community. Assuming that this makes it easier for reviewers to judge these type of submissions, researchers might feel more confident in submitting such papers. Alternatively, authors could be given the opportunity to specifically protest reviewers’ decisions (ex post), if they believe their method is of relevance and the negative results overly contributed to a rejection. For the latter example, measures would have to be taken to avoid inflationary use of such an option. 
Notably, these mechanisms are already in place to some degree:
If a new method is proposed in an NMNR-publication, it is the authors' responsibility to make a case for the worth of their work. Similarly, a rebuttal phase, which is an integral part of the review process for several venues, presents opportunities to address perceived unfair judgment.
However, given the overemphasis on positive results, we believe a conscious inclusion of the described elements ex ante or ex post could be beneficial to shift the paradigm regarding negative results.

While our proposed ideas are not intended as finished ``plug-and-play''-solutions, we hope they can serve as a good starting point for the community to discuss if and how the review process should be adapted to better accommodate negative results.

\subsection{Emphasize certain criteria when assessing the impact of papers with negative results during the review process.}
\label{ssec:review-criteria-negative-results}
We have thus far expressed that reviewers are a critical piece of the puzzle when it comes to normalizing negative results.
We have also stated that the review process provides very reasonable guidelines to judge the merit of a submission, but that these guidelines can be quite abstract.
We therefore present several more concrete criteria that could be derived from abstract notions like soundness or significance~\cite{icml-review-form}.
To better showcase this in the context of negative results, we provide examples for criteria that are of equal importance for positive and negative results, criteria that are especially important in the context of NMNR-publications and criteria that are especially important in the context of EMNR-publications.
We want to emphasize that this is not a perfect or comprehensive list and we do not want to present it as such. However, as this position aims at starting a discussion, we again want to provide a concrete starting point. \medskip\newline
\textbf{1) Criteria relevant to both positive and negative results:}

\textbf{An open, well written and easy-to-use codebase.} This is certainly desirable for any publication proposing a novel method, but arguably even more important for negative results. Reproducibility and possibility for future work should be valued highly for both negative and positive results, but a strong argument for the publication of negative results is often the opportunities that stem from other researchers basing future work on them. Additionally, if the negative results are especially surprising, a meticulous setup including accessible code is even more so important, because it needs to be clear that these surprising results are not due to poor implementation.
We therefore believe that an open, well written and easy-to-use codebase is critical for all empirical machine learning publications, but even more important in the context of negative results (both for NMNR- and EMNR-publications).

\textbf{Experimental design and setup.} Proper and meticulous experimentation is an important foundation for all conclusions drawn in empirical machine learning papers. We believe this to be equally important for all positive and negative results and the criterion should be interpreted and executed in a similar fashion for both types.

\textbf{2) Criteria specifically aimed at NMNR-publications:}

\textbf{Surprise factor and “obviousness” of negative results.} While a surprising result is generally regarded as positive, because it implies novelty, this is not a crucial factor for many papers with positive results. After all, if a new method is proposed and it clearly shows an improvement over the state-of-the-art, a surprise factor is not necessary to make this an impactful contribution. If, on the other hand, a negative result is very predictable and obvious, there really is not much sense in publishing it. One could think of some extreme cases that serve as counterexamples, like e.g., a certain negative result that can only be demonstrated with immense effort that has not been conclusively shown because of this, but in general this holds true. Algorithmic novelty is strongly correlated with this criterion, but could be considered as an additional or alternative criterion.

\textbf{Depth of analysis of negative results.} In NMNR-publications it can be especially important to also foster an understanding of the observed results. Ideally, authors would provide an in-depth analysis as to why the negative results were observed as there is a lot of value in this for the community. If this is not possible, authors should provide an intuition as well as research questions to the community that could help obtain this understanding. If this is also not possible, the authors should at least give a commentary as to if the further exploration of this understanding is a worthwhile endeavor for the community. If they deem it so, they should additionally explain why they are unable to provide it and unable to provide related research questions.

\textbf{3) Criteria specifically aimed at EMNR-publications:}

\textbf{Ethical considerations and societal implications due to newly discovered failure modes.} While ethical implications should always be considered in research, some specific questions arise for EMNR-publications, since often state-of-the-art methods are examined. Established methods are likely heavily used in a variety of applications. Will the newly discovered failure modes endanger any important applications? How can this be combated moving forward? Can the newly discovered failure mode be shored up and thus make some high-risk applications more robust and reliable?

\textbf{Depth of analysis of negative results.} Similarly to NMNR-publications the understanding and analysis of failure modes and other results is very important in the context of EMNR-publications. The more detailed description from above carries over to this point.

\section{Conclusion}
Empirical machine learning publications that propose novel methods are often mainly judged by their exhibited predictive performance on a problem set chosen by the authors.
As a result, most published papers of this type feature impressive results and claim to beat the current state-of-the-art.
In this position paper, we put out a call to the research community for a paradigm shift towards normalizing the publication of negative results.
We provide a detailed analysis on why neglecting negative results is problematic---especially when mainly using predictive performance for assessing the value of a contribution to our research field. 

We further outline the advantages of publishing negative results and how this can improve machine learning research with respect to efficiency, practical relevance, diversity and overall advancement of the research field.
The paper concludes with proposing eight concrete action points that can be implemented to help the machine learning research community move towards this new paradigm.

\section*{Acknowledgements}
We thank our colleagues for several interesting discussions centered around negative results and would like to especially thank Thomas Seidl on productive discussion on what actually constitutes a negative result in empirical machine learning.

The authors acknowledge support by the Bavarian Ministry of Economic Affairs, Regional Development and Energy through the Center for Analytics – Data – Applications (ADA-Center) within the framework of BAYERN DIGITAL II (20-3410-2-9-8).

\section*{Impact Statement}
This paper presents a position and deliberation on the role of negative results in empirical machine learning research. Its goal is to spark discussion in the machine learning community and overall improve and further the field of machine learning. There are many potential societal consequences of our work, none which we feel must be specifically highlighted here.


\bibliography{bibliography}
\bibliographystyle{icml2024}

\newpage



\end{document}